\documentclass[10pt]{article}

\newcommand{\papertitle}{Benchmarking Frontier Text-to-Image Models on the Image Description Prompts}
\newcommand{\papersubtitle}{An Independent-Judge Rubric Evaluation of Four Production Image-Generation Systems}
\newcommand{\shorttitle}{Frontier Text-to-Image Benchmark on Image Description Prompts}
\newcommand{\organizationname}{Perle}
\newcommand{\paperdate}{\today}
\newcommand{\logofile}{company_logo}

\usepackage{iftex}
\ifPDFTeX
  \usepackage[utf8]{inputenc}
  \usepackage[T1]{fontenc}
  \usepackage{lmodern}
\else
  \usepackage{fontspec}
\fi

\usepackage{microtype}
\usepackage[
  top=2.5cm,
  bottom=2.8cm,
  left=2.5cm,
  right=2.5cm
]{geometry}
\usepackage{amsmath}
\usepackage{amssymb}
\usepackage{graphicx}
\usepackage{float}
\usepackage[
  labelfont=bf,
  font=small,
  justification=justified,
  singlelinecheck=false
]{caption}
\usepackage{booktabs}
\usepackage{array}
\usepackage{ragged2e}
\usepackage[table]{xcolor}
\usepackage{tabularx}
\usepackage{enumitem}
\usepackage{fancyhdr}
\usepackage{titlesec}
\usepackage{longtable}
\usepackage{multirow}
\usepackage{xurl}
\usepackage{tikz}
\usetikzlibrary{shapes.geometric, arrows.meta, positioning, fit, backgrounds}
\usepackage{pifont}
\usepackage[
  colorlinks=true,
  linkcolor=black,
  citecolor=black,
  urlcolor=black
]{hyperref}

\newcolumntype{C}{>{\centering\arraybackslash}X}
\newcolumntype{L}{>{\raggedright\arraybackslash}X}
\newcolumntype{P}[1]{>{\RaggedRight\arraybackslash}p{#1}}

\DeclareRobustCommand{\modelname}[1]{%
  \texttt{\hyphenchar\font=45\relax #1}}
\DeclareRobustCommand{\emailaddr}[1]{%
  \href{mailto:#1}{\nolinkurl{#1}}}

\definecolor{brandcolor}{HTML}{000000}
\definecolor{lightlinecolor}{HTML}{CCCCCC}
\definecolor{rowgray}{HTML}{F5F5F5}

\titleformat{\section}
  {\normalfont\large\bfseries}{\thesection.}{0.5em}{}
\titlespacing{\section}{0pt}{18pt plus 3pt minus 2pt}{6pt}

\titleformat{\subsection}
  {\normalfont\normalsize\bfseries}{\thesubsection.}{0.5em}{}
\titlespacing{\subsection}{0pt}{12pt plus 2pt minus 1pt}{4pt}

\titleformat{\subsubsection}[runin]
  {\normalfont\normalsize\bfseries}{\thesubsubsection.}{0.5em}{}[.\quad]
\titlespacing{\subsubsection}{0pt}{8pt plus 1pt}{0pt}

\setlist[itemize]{leftmargin=1.5em, topsep=3pt, itemsep=2pt, parsep=0pt}
\setlist[enumerate]{leftmargin=1.5em, topsep=3pt, itemsep=2pt, parsep=0pt}

\fancypagestyle{firstpage}{
  \fancyhf{}
  
  \fancyfoot[L]{\footnotesize
    \textcopyright~\the\year~\organizationname. All rights reserved.}
  \fancyfoot[R]{\small 1}
}

\renewenvironment{abstract}{%
  \small\bfseries\noindent\ignorespaces
}{\par\medskip}

\newcommand{\brandrule}{%
  \noindent\textcolor{brandcolor}{\rule{\linewidth}{1.5pt}}\par}
\newcommand{\lightrule}{%
  \noindent\textcolor{lightlinecolor}{\rule{\linewidth}{0.4pt}}\par}

\begin{document}
\thispagestyle{firstpage}

\begin{flushleft}
  \IfFileExists{\logofile.png}{%
    \includegraphics[height=20pt]{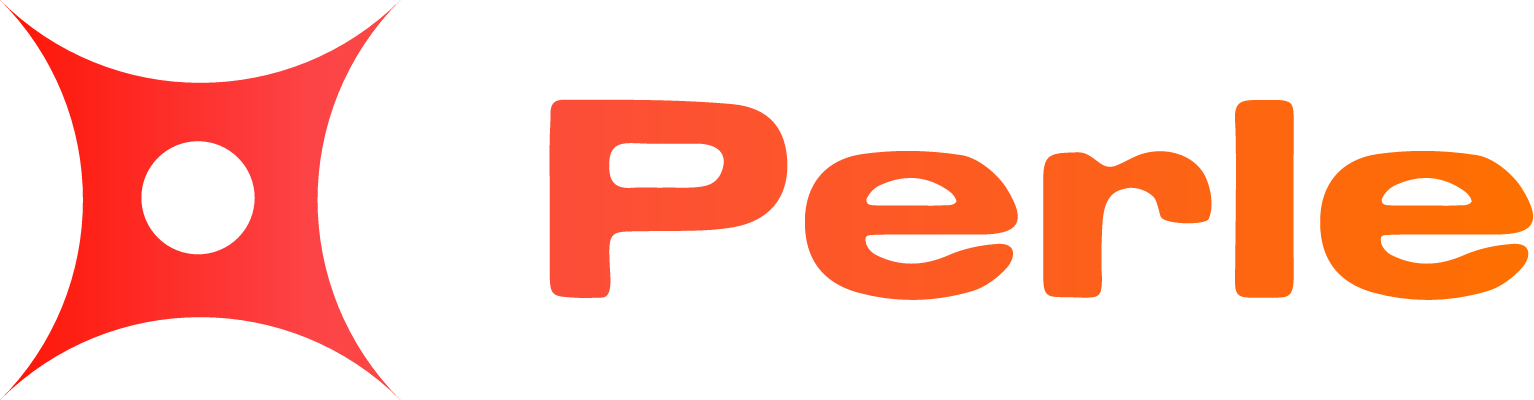}%
  }{%
    \textbf{\organizationname}%
  }
  \hfill
  {\small\paperdate}
\end{flushleft}

\vspace{4pt}
\brandrule
\vspace{8pt}

\noindent{\fontsize{17pt}{20pt}\selectfont\bfseries
  \papertitle\\[3pt]
  \large\papersubtitle\par}

\vspace{10pt}

\noindent
\textbf{Sajjad Abdoli}\textsuperscript{1,*,\dag}\footnote{Corresponding author: \emailaddr{sajjad@perle.ai}},\allowbreak\quad
\textbf{Ghassan Al-Sumaidaee}\textsuperscript{1,*,\dag}\footnote{Corresponding author: \emailaddr{ghassan.al-sumaidaee@perle.ai}; ORCID: \href{https://orcid.org/0000-0002-5536-0252}{0000-0002-5536-0252}},\allowbreak\quad
\textbf{Ahmed Rashad}\textsuperscript{1}
\par\vspace{3pt}
{\small\textsuperscript{1}Perle\quad
\textsuperscript{*}Equal contribution; names sorted alphabetically.\quad
\textsuperscript{\dag}Corresponding authors\\[2pt]
\emailaddr{sajjad@perle.ai}\quad
\emailaddr{ghassan.al-sumaidaee@perle.ai}\\[2pt]
\emailaddr{ahmed@perle.ai}}

\vspace{10pt}

\begin{abstract}
Text-to-image models are typically reported on average-case prompts, which understates the gap
between systems on the compositionally demanding requests real users actually issue -- precise
object counts, multi-object attribute binding, legible embedded text, and explicit spatial
constraints. We evaluate four production text-to-image systems -- Hunyuan 3.0, Gemini 3 Pro Image
(``Nano Banana Pro''), Black Forest Labs FLUX.2, and Ideogram 3.0 -- on the 48 hardest prompts
drawn from the DataSeeds.AI Sample Dataset (DSD), selected by an automated complexity-scoring pass
over the full corpus. Every generated image is graded by an independent-judge rubric: one model
(GPT-5.4-Pro) authors an atomic, weighted, Mutually Exclusive and Collectively Exhaustive (MECE) evaluation rubric for each generated image, and a
second, different model (Gemini 3.1 Pro Preview) independently renders every met/triggered verdict
against that rubric -- so the model that decides what would count as satisfying the prompt is never
the model that decides whether a given image actually does. On this matched set of 48
identical prompts, Gemini 3 Pro Image ranks first (84.8/100), narrowly ahead of FLUX.2 (82.3/100),
with Ideogram 3.0 (65.7/100) and Hunyuan 3.0 (63.3/100) trailing by a wide margin. Failure-code
analysis shows the two leading systems mainly lose points to miscounted objects and geometric
artifacts, while the trailing systems more frequently garble rendered text
(\texttt{TEXT\_GARBLING}) with Ideogram 3.0 in particular omitting requested elements entirely (\texttt{MISSING\_STEP}). The full per-sample rubric, scores, and failure annotations for all 48 prompts across all four models are available from the authors on request.
\end{abstract}

\noindent\textbf{Keywords:} text-to-image generation; benchmark evaluation; LLM-as-judge; rubric-based grading; compositional prompting; DataSeeds.AI DSD

\vspace{6pt}
\lightrule
\vspace{14pt}

\section{Introduction}
\label{sec:introduction}

Frontier text-to-image (T2I) systems are improving quickly on aggregate benchmark scores, but
aggregate scores are dominated by simple, low-constraint prompts on which most modern models
already perform well. The prompts that best separate model quality in practice are the
compositionally demanding ones: precise object counts, several simultaneous attribute bindings,
explicit spatial relations, legible embedded text, and physics-dependent visual elements such as
reflections and shadows \cite{huang2023t2icompbench, saharia2022imagen}. A model comparison that
does not deliberately select for this kind of difficulty risks concluding that several systems are
roughly equivalent, when in fact they diverge sharply once compositional load increases.

This paper reports a focused, difficulty-targeted comparison of four production T2I systems on the
48 hardest prompts in the DataSeeds.AI Sample Dataset (DSD), identified by an automated
complexity-scoring pass over the dataset rather than by manual curation. Every generated image is
graded by an independent-judge rubric methodology in which the model that authors the grading
rubric is never the model that applies it, removing the self-grading bias that affects
single-model-as-judge setups.

Our contributions are:

\begin{itemize}
  \item A reproducible, automated method for selecting the highest-complexity prompts out of a
  large prompt corpus, using an LLM complexity classifier rather than manual selection.
  \item An independent-judge, atomic/weighted rubric methodology for grading T2I outputs, with a
  writer model and a scorer model that are never the same, and a fixed 11-code MECE failure
  taxonomy covering both missed requirements and actively introduced errors.
  \item A matched-N, same-prompt comparison of four frontier T2I systems (Hunyuan 3.0, Gemini 3 Pro
  Image, FLUX.2, Ideogram 3.0) on the 48 hardest DSD prompts, with per-model failure-code
  breakdowns that explain \emph{why} each system loses points, not just its aggregate score.
\end{itemize}

\section{Background and Related Work}
\label{sec:background}

\subsection{How Text-to-Image Models Work}
\label{sec:t2i-background}

Modern text-to-image systems share a common shape: a text encoder converts the prompt into a
sequence of embeddings, and a generative backbone conditions on those embeddings to produce an
image, typically in a compressed latent space rather than directly in pixels. Three backbone
families are represented among the systems compared in this paper. \emph{Diffusion} models learn to
reverse a gradual noising process, iteratively denoising a latent conditioned on the text embedding
at each step \cite{saharia2022imagen}. \emph{Autoregressive} models instead generate an image as a
sequence of discrete visual tokens, predicted one at a time much like a language model predicts the
next word \cite{yu2022parti}; Hunyuan Image 3.0 \cite{tencent2025hunyuanimage3} is a
large-scale example of this family, unifying understanding and generation in a single
Mixture-of-Experts autoregressive model. \emph{Rectified-flow} (flow-matching) models, used by
FLUX.2 \cite{bfl2025flux2}, learn a velocity field that transports noise to data along a straighter
path than classical diffusion, which can reduce the number of sampling steps needed for a given
quality level. Regardless of backbone, none of these architectures explicitly represents discrete,
symbolic constraints such as ``exactly four objects'' or ``the sign reads OPEN'' -- every constraint
must instead emerge from a continuous, learned mapping between text embeddings and pixels. This is
the underlying reason precise counting, exact text rendering, and multi-object attribute binding
remain comparatively hard for every system in this study regardless of scale or architecture: these
requirements are trivial to \emph{state} in a prompt but are not natively represented in how the
model produces an image, unlike, say, overall color palette or scene composition, which map more
directly onto the smooth, continuous features these models learn well.

\subsection{Compositional Text-to-Image Benchmarks}

Prior T2I evaluation suites such as DrawBench \cite{saharia2022imagen} and PartiPrompts
\cite{yu2022parti} introduced curated prompt sets spanning categories like counting, spatial
relations, and text rendering, and showed that model rankings can change substantially depending
on which prompt category is emphasized. T2I-CompBench \cite{huang2023t2icompbench}, later extended
as T2I-CompBench++ \cite{huang2025t2icompbenchpp}, formalized compositional evaluation further,
decomposing performance into attribute binding, object relationships, and complex compositions.
HEIM \cite{lee2023holistic} broadened evaluation to a holistic set of axes including alignment,
aesthetics, toxicity, and bias. More recently, R2I-Bench \cite{chen2025r2ibench} extended
compositional evaluation into reasoning-driven prompts (commonsense, mathematical, logical, and
causal constraints that must be inferred rather than read off literally), and FineGRAIN
\cite{hayes2025finegrain} introduced a structured, 27-category failure-mode taxonomy scored by
vision-language-model judges -- closely related in spirit to the failure taxonomy used in this
paper (Table~\ref{tab:taxonomy}), though FineGRAIN targets open research checkpoints rather than
production/API-served commercial systems. Our work differs from all of the above by selecting hard
prompts \emph{automatically} from a single existing photograph-captioning dataset via a learned
complexity classifier rather than from a purpose-built prompt suite, and by using a full
independent-judge rubric authored fresh for every individual prompt rather than a fixed,
prompt-set-wide metric or a shared static taxonomy applied uniformly across prompts.

\subsection{LLM-as-Judge and Self-Grading Bias}

Using a large language or vision-language model to grade another model's output is now common
practice, but a judge model grading outputs from a model family it is closely related to (or
grading a rubric it wrote itself) can introduce systematic bias. We address this directly by
splitting rubric authorship and rubric application across two different frontier models, so the
model that decides what counts as correct is never the model that decides whether a given image
satisfies it.

\section{Data}
\label{sec:data}

Prompts are drawn from the DataSeeds.AI Sample Dataset (DSD) \cite{dataseeds2024}, a corpus of
real-world photographs each paired with a human-written \texttt{image\_description} and
\texttt{scene\_description}. For every sample, the generation prompt is the verbatim concatenation
of these two fields -- no paraphrasing, truncation, or manual editing is applied at any stage.

\subsection{Complexity Scoring and Hardest-Prompt Selection}

Of the roughly 7{,}009 valid rows in DSD (rows with both text fields populated), every candidate
prompt was scored for generation difficulty by an LLM complexity classifier (Gemini 3.1
Flash-Lite) on a 1--10 scale, with the rubric shown in Table~\ref{tab:complexity-rubric}. The
classifier also returns a short list of qualitative \emph{hardness factors} for each prompt (e.g.
``precise text rendering'', ``multiple specific text elements'', ``nested structural
requirements'') alongside the numeric score.

\begin{table}[H]
\centering
\caption{Complexity/hardness scoring rubric applied to every candidate DSD prompt.}
\label{tab:complexity-rubric}
\small
\begin{tabularx}{\textwidth}{L L}
  \toprule
  \textbf{Score range} & \textbf{Definition} \\
  \midrule
  \rowcolor{rowgray} 1--3 & Simple single-subject scene, no counting, no exact spatial or attribute constraints. \\
  4--6 & Moderate: a few objects, some attribute or spatial relations, mild counting. \\
  \rowcolor{rowgray} 7--10 & Hard: precise counts of multiple object classes, many simultaneous attribute bindings, tight spatial/compositional constraints, physics-dependent elements (reflections, shadows, transparency), legible text/script, or many named entities that must each be individually correct. \\
  \bottomrule
\end{tabularx}
\end{table}

We selected the \textbf{48 hardest prompts} in the dataset by this complexity score, bounding the
study to this size given the cost of running independent-judge grading (Section~\ref{sec:methodology})
across four production models. Because a large number of prompts cluster at the top of the 1--10
scale, ties at the maximum observed score were broken with a seeded random shuffle prior to a
stable sort, so that prompts appearing later in the raw dataset are not systematically favored or
disfavored at the cutoff. Table~\ref{tab:data-summary} summarizes the resulting evaluation set.

\begin{table}[H]
\centering
\caption{Evaluation set summary. Complexity is scored 1--10 (higher = harder); the selected set sits at the top of the observed distribution.}
\label{tab:data-summary}
\small
\begin{tabularx}{0.8\textwidth}{L C}
  \toprule
  \textbf{Property} & \textbf{Value} \\
  \midrule
  \rowcolor{rowgray} Source dataset & DataSeeds.AI Sample Dataset (DSD) \\
  Candidate pool scored for complexity & 7{,}009 valid rows \\
  \rowcolor{rowgray} Selected evaluation prompts & 48 (hardest by complexity score) \\
  Complexity score of selected prompts & 9/10 (maximum observed) \\
  \rowcolor{rowgray} Models compared & 4 (Hunyuan 3.0, Gemini 3 Pro Image, FLUX.2, Ideogram 3.0) \\
  Images graded per model & 48 \\
  \rowcolor{rowgray} Total graded (image, rubric) pairs & 192 \\
  \bottomrule
\end{tabularx}
\end{table}

Recurring hardness factors among the selected prompts include: multiple legible text elements
that must render correctly (often with an exact string to match), precise counts of several named
object classes simultaneously, specific spatial arrangements among three or more objects, and
attribute bindings (color, material) that must attach to the correct object rather than bleed to a
neighboring one.

\section{Methodology}
\label{sec:methodology}

\subsection{Pipeline Overview}

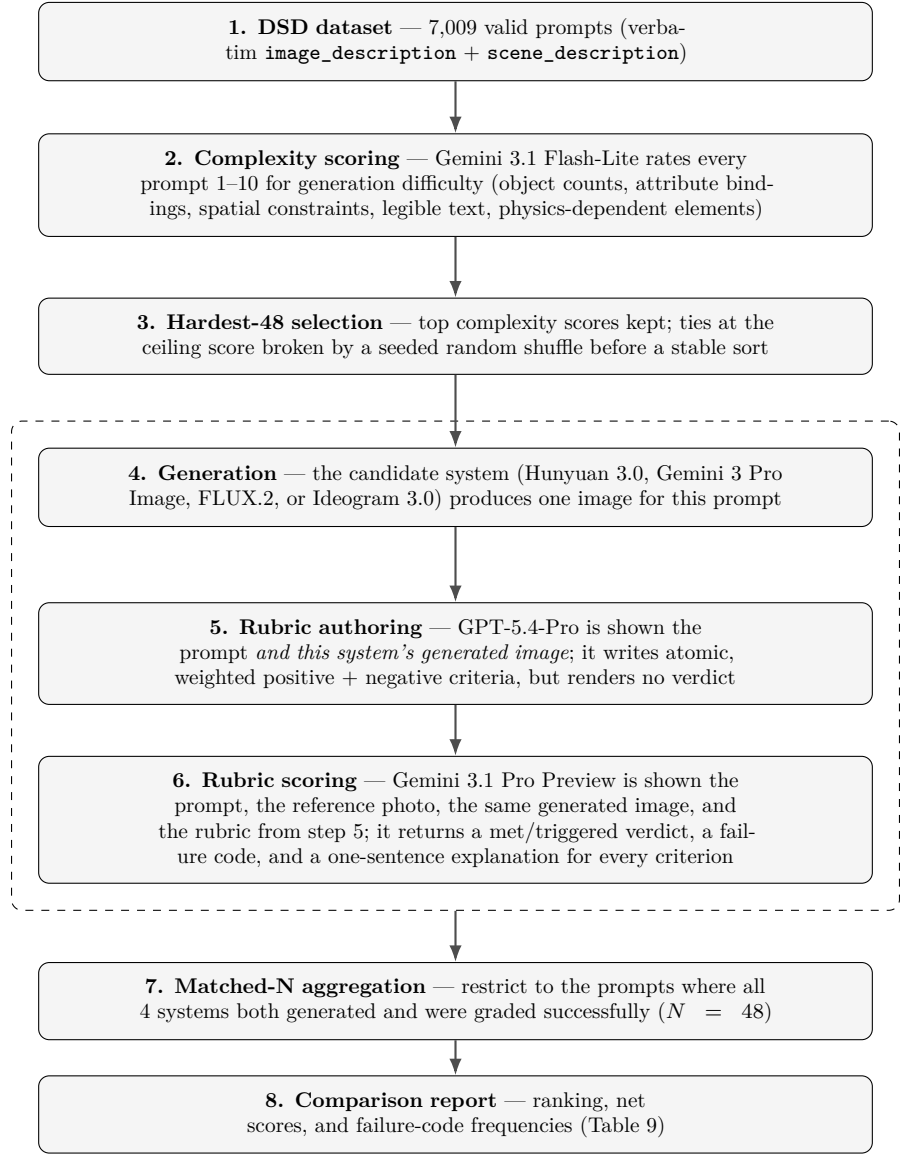
\begin{figure}[H]
\centering
\begin{tikzpicture}[
  every node/.style={font=\footnotesize},
  box/.style={draw, rounded corners, align=center, fill=rowgray, inner sep=6pt, text width=10.6cm},
  arrow/.style={-{Latex[length=2.2mm]}, thick, draw=black!70}
]

\node[box] (dsd) at (7,17.6) {\textbf{1. DSD dataset} --- 7{,}009 valid prompts (verbatim \texttt{image\_description} + \texttt{scene\_description})};
\node[box] (complexity) at (7,15.7) {\textbf{2. Complexity scoring} --- Gemini 3.1 Flash-Lite rates every prompt 1--10 for generation difficulty (object counts, attribute bindings, spatial constraints, legible text, physics-dependent elements)};
\node[box] (select) at (7,13.7) {\textbf{3. Hardest-48 selection} --- top complexity scores kept; ties at the ceiling score broken by a seeded random shuffle before a stable sort};
\node[box] (gen) at (7,11.7) {\textbf{4. Generation} --- the candidate system (Hunyuan 3.0, Gemini 3 Pro Image, FLUX.2, or Ideogram 3.0) produces one image for this prompt};
\node[box] (rubric) at (7,9.5) {\textbf{5. Rubric authoring} --- GPT-5.4-Pro is shown the prompt \emph{and this system's generated image}; it writes atomic, weighted positive + negative criteria, but renders no verdict};
\node[box] (score) at (7,7.3) {\textbf{6. Rubric scoring} --- Gemini 3.1 Pro Preview is shown the prompt, the reference photo, the same generated image, and the rubric from step 5; it returns a met/triggered verdict, a failure code, and a one-sentence explanation for every criterion};

\node[draw, dashed, rounded corners, fit=(gen)(rubric)(score), inner sep=10pt] (repeatbox) {};

\node[box] (agg) at (7,4.9) {\textbf{7. Matched-N aggregation} --- restrict to the prompts where all 4 systems both generated and were graded successfully ($N=48$)};
\node[box] (report) at (7,3.4) {\textbf{8. Comparison report} --- ranking, net scores, and failure-code frequencies (Table~\ref{tab:main-results})};

\draw[arrow] (dsd) -- (complexity);
\draw[arrow] (complexity) -- (select);
\draw[arrow] (select) -- (gen);
\draw[arrow] (gen) -- (rubric);
\draw[arrow] (rubric) -- (score);
\draw[arrow] (repeatbox.south) -- (agg.north);
\draw[arrow] (agg) -- (report);

\end{tikzpicture}
\caption{End-to-end benchmarking pipeline, from the raw DSD dataset to the final comparison
report. The dashed box (steps 4--6) is repeated independently for each of the 4 candidate systems.
Steps 4--6 are \emph{sequential}, not parallel, within that box: rubric authoring (step 5) depends
on the specific image produced in step 4, since the writer is shown that image. The independence
that matters for avoiding self-grading bias is not ``the writer never sees an image'' but ``the
writer never renders a verdict on its own rubric'' -- that binary judgment is made independently in
step 6 by a different model.}
\label{fig:pipeline}
\end{figure}

Figure~\ref{fig:pipeline} summarizes the full pipeline from raw dataset to comparison report.
Prompt selection (steps 1--3) is fully automated and independent of any model's generation
quality. Steps 4--6 then run once per candidate system, on that system's own generated image: the
rubric writer is deliberately shown the specific image it is writing criteria for (so it can judge
which failure categories are actually plausible), but a \emph{different} model renders the
met/triggered verdicts in step 6, so the writer never judges its own rubric.

\subsection{Study Design}

Each of the 48 prompts is submitted, verbatim and identically, to all four candidate systems --
Hunyuan Image 3.0 \cite{tencent2025hunyuanimage3}, Gemini 3 Pro Image
\cite{deepmind2026gemini3proimage}, FLUX.2 \cite{bfl2025flux2}, and Ideogram 3.0
\cite{ideogram2025v3} (Table~\ref{tab:models}) -- producing one generated image per (prompt, model)
pair. Exact generation parameters for each system are listed in Appendix~\ref{app:gen-params}.

\begin{table}[H]
\centering
\caption{Candidate text-to-image systems and generation route.}
\label{tab:models}
\small
\begin{tabularx}{\textwidth}{L L}
  \toprule
  \textbf{System} & \textbf{Generation route} \\
  \midrule
  \rowcolor{rowgray} Hunyuan 3.0 \cite{tencent2025hunyuanimage3} & \modelname{hunyuan-image} endpoint (Pixazo) \\
  Gemini 3 Pro Image \cite{deepmind2026gemini3proimage} (``Nano Banana Pro'') & Direct via Google GenAI (\modelname{gemini-3-pro-image}) \\
  \rowcolor{rowgray} Black Forest Labs FLUX.2 \cite{bfl2025flux2} & Direct via Black Forest Labs (\modelname{flux-2-pro-preview}) \\
  Ideogram 3.0 \cite{ideogram2025v3} & Direct via Ideogram (\modelname{ideogram-v3/generate}) \\
  \bottomrule
\end{tabularx}
\end{table}

\subsection{Independent-Judge Rubric Grading}

Every generated image is graded through a two-stage, two-model process designed so that no single
model ever both authors and applies its own grading criteria.

\subsubsection{Rubric authoring}

For each (prompt, model) pair, a writer model (GPT-5.4-Pro, called directly against OpenAI's
Responses API) is shown \emph{both} the text prompt \emph{and} the specific image that system
generated from it, and produces two atomic, weighted criteria sets. Critically, the writer is
instructed to output only the criteria and their weights -- \emph{not} a verdict on whether any
criterion is satisfied; that judgment is withheld entirely for the separate scoring step
(Section~4.3.2), carried out by a different model:

\begin{itemize}
  \item \textbf{Positive criteria} (4--8 per prompt): atomic, self-contained, MECE claims derived from the prompt's stated requirements (object counts, colors, spatial relations, actions, explicit constraints), whose set and weights are the same across systems for a given prompt even though wording may be grounded in the observed image -- each weighted $+1$ to $+10$ by how essential it is to a minimally acceptable
  response (core requirements score higher than nice-to-have enhancements).
  \item \textbf{Negative criteria}: atomic claims about failure modes the model might introduce on
  its own initiative, independent of what the prompt asked for -- anatomical/geometric artifacts,
  hallucinated extraneous objects, illegible text, physically implausible lighting/shadows, color
  bleeding between adjacent objects, and violations of any explicit negative constraint in the
  prompt -- each weighted $-1$ to $-10$ by severity.
\end{itemize}

\paragraph{Why the writer sees the image, and why rubrics differ by model.} The writer is given a
fixed, pre-defined checklist of seven generic failure-mode categories known a priori to be common
text-to-image pathologies -- geometric/anatomical artifacts, hallucinated objects, duplicated
subjects, illegible text, implausible lighting/reflections, color bleeding, and explicit
negative-constraint violations -- and, \emph{for every image}, considers each category and includes
one atomic criterion per category that is plausible for \emph{this specific image} (skipping a
category only if clearly inapplicable, e.g.\ no text-legibility criterion if no text appears
anywhere in the image or prompt). Seeing the actual image lets the writer ground each criterion in
concrete, relevant visual content rather than guessing generically from the prompt text alone --
for instance, it can write a duplication criterion that names the specific object type it observes
repeated, rather than a vaguer catch-all. This is also the direct reason the rubric is authored
\emph{independently for each (prompt, model) pair} rather than once per prompt and reused across all
four systems: because each system produces a different image for the same prompt, the writer is
shown different visual content each time and tailors which failure categories are applicable (and
how each criterion is worded) accordingly, even though the underlying prompt-derived positive
criteria target the same requirements across models. The writer is never told which system
generated the image, and -- critically -- never renders a met/triggered verdict itself; that
judgment is made independently, by a different model, in the next step. This keeps the two roles
genuinely separate: the writer decides what would count as satisfying the prompt (and what could
plausibly go wrong), while a different model alone decides whether this particular image actually
does.

Every criterion is required to be self-contained (stating the expected value inline, e.g. ``shows
exactly 3 balloons'' rather than ``shows the correct number of balloons'') and specific enough that
independent graders would be expected to agree on the verdict at least 75\% of the time. Criteria
are never phrased as the logical negation of one another, so a single fact is never scored twice.

\subsubsection{Rubric scoring}

A separate scorer model (Gemini 3.1 Pro Preview) is then shown the original prompt, the reference
photograph, the generated image, and the rubric produced above -- but never the identity of the
generating model or of the rubric author. For every criterion it returns a binary verdict (met /
not met for positive criteria; triggered / not triggered for negative criteria) together with a
failure code drawn from a fixed taxonomy (Table~\ref{tab:taxonomy}) and a one-sentence,
evidence-citing explanation.

\begin{table}[H]
\centering
\caption{Fixed MECE failure taxonomy used for every criterion verdict.}
\label{tab:taxonomy}
\small
\begin{tabularx}{\textwidth}{L L}
  \toprule
  \textbf{Code} & \textbf{Meaning} \\
  \midrule
  \rowcolor{rowgray} \texttt{MISSING\_STEP} & A requested element is absent from the output. \\
  \texttt{HALLUCINATED\_OBSERVATION} & An unrequested element was invented. \\
  \rowcolor{rowgray} \texttt{FORMAT\_VIOLATION} & Wrong size/ratio/output type. \\
  \texttt{CONSTRAINT\_BREACH} & An explicit non-numeric, non-spatial constraint was broken. \\
  \rowcolor{rowgray} \texttt{INCOMPLETE\_TRAJECTORY} & Output is cropped, truncated, or partial. \\
  \texttt{MISCOUNT} & Wrong number of some countable object/element. \\
  \rowcolor{rowgray} \texttt{TEXT\_GARBLING} & Illegible or incorrect rendered text/glyphs. \\
  \texttt{PHYSICS\_VIOLATION} & Impossible shadows, gravity, reflections, occlusion, transparency. \\
  \rowcolor{rowgray} \texttt{SPATIAL\_RELATION\_ERROR} & Left/right/above/below/behind/in-front relation is wrong. \\
  \texttt{ATTRIBUTE\_BINDING\_ERROR} & A color/material/size attribute is attached to the wrong object. \\
  \rowcolor{rowgray} \texttt{GEOMETRY\_ARTIFACT} & Fused/melted parts, extra or missing limbs, non-manifold shapes. \\
  \bottomrule
\end{tabularx}
\end{table}

\subsection{Evaluation Metrics}

Each (prompt, model) evaluation yields a single normalized quality score in $[0, 100]$:

\begin{align}
  \text{final\_score} = \max\!\left(0,\ \frac{W^{+}_{\text{earned}} + W^{-}_{\text{incurred}}}{W^{+}_{\text{possible}}} \times 100\right)
\end{align}

where $W^{+}_{\text{earned}}$ is the sum of weights for positive criteria judged \emph{met},
$W^{-}_{\text{incurred}}$ is the (already negative) sum of weights for negative criteria judged
\emph{triggered}, and $W^{+}_{\text{possible}}$ is the sum of \emph{all} positive-criteria weights
for that prompt. Dividing by the prompt-specific total possible weight normalizes away the fact
that harder prompts naturally carry more total positive weight than simpler ones, so scores stay
comparable across prompts of different compositional load. The score is floored at 0 so that a
model which accumulates more negative penalty than positive credit reads as 0/100 rather than a
confusing negative percentage.

We additionally report, per model: \emph{net score} (mean of $W^{+}_{\text{earned}} +
W^{-}_{\text{incurred}}$, unnormalized), \emph{average unmet criteria} (mean count of positive
criteria judged not met), and \emph{average active errors} (mean count of negative criteria judged
triggered).

\section{Worked Rubric Examples}
\label{sec:worked-examples}

Aggregate scores alone do not convey \emph{why} one model outscores another. This section walks
through two prompts in full detail -- showing the actual rubric criteria, weights, and per-criterion
verdicts -- so the scoring mechanism in Section~\ref{sec:methodology} is concrete rather than
abstract. Recall that the rubric is authored independently for each (prompt, model) pair, with the
writer shown that specific model's own generated image (Section~4.3.1); so criterion wording and
which negative-criteria categories are included vary by model, tailored to what the writer actually
observed in each image, even though the underlying prompt-derived requirements are the same. Final
scores remain directly comparable across models because Equation~1 normalizes by each rubric's own
total possible weight.

\subsection{Example: ``Car-shaped Mailbox''}

\textbf{Prompt:} \textit{``This image features a car-shaped mailbox, with its sign displaying the
address: `1118 PRINTERY RD.' The road is covered with snow on both sides, with blurred buildings on
the background right side. [...] The mailbox has four large black and yellow tires attached to a
green metal frame. [...]''}

\begin{figure}[H]
  \centering
  \IfFileExists{figures/example_mailbox.png}{%
    \includegraphics[width=\textwidth]{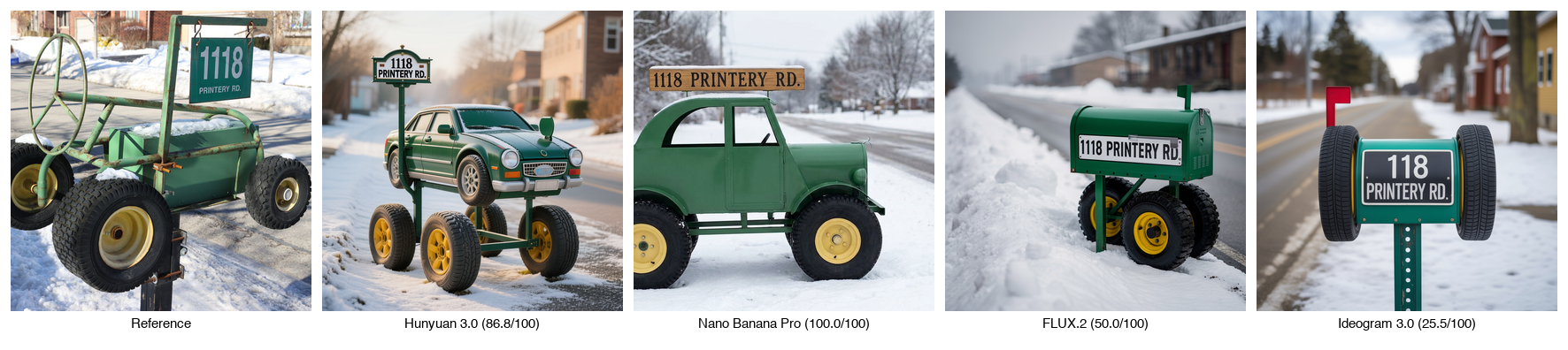}%
  }{%
    \fbox{\parbox[c][3.5cm][c]{0.95\textwidth}{\centering
      Replace this box with \texttt{figures/example\_mailbox.png}.}}%
  }
  \caption{Reference photograph and all four generated outputs for this prompt. Every system renders
  the exact address string correctly; the differentiating failure is whether the object reads as
  ``car-shaped'' at all (Gemini 3 Pro Image, Hunyuan 3.0) versus a generic wheeled mailbox (FLUX.2,
  Ideogram 3.0), despite FLUX.2 still outscoring Ideogram 3.0 overall on other criteria.}
  \label{fig:example-mailbox}
\end{figure}

Table~\ref{tab:example1-rubric} shows the complete rubric authored for FLUX.2's output on this
prompt, together with the scoring model's verdict on every criterion.

\begin{table}[H]
\centering
\caption{Full rubric and verdicts for FLUX.2 on the ``Car-shaped Mailbox'' prompt (final score 50.0/100). \checkmark = met/not
triggered; \ding{55} = not met/triggered (weight then counts against the model).}
\label{tab:example1-rubric}
\small
\begin{tabularx}{\textwidth}{P{8.7cm} C C L}
  \toprule
  \textbf{Criterion} & \textbf{Weight} & \textbf{Verdict} & \textbf{Failure code} \\
  \midrule
  \multicolumn{4}{l}{\textit{Positive criteria (possible weight: 54)}} \\
  \rowcolor{rowgray} The primary subject is a mailbox designed to resemble a car. & +10 & \ding{55} & \texttt{MISSING\_STEP} \\
  The mailbox body is green. & +7 & \checkmark & -- \\
  \rowcolor{rowgray} A sign on the mailbox reads exactly ``1118 PRINTERY RD.'' & +10 & \checkmark & -- \\
  A green metal frame is visible beneath the mailbox. & +5 & \checkmark & -- \\
  \rowcolor{rowgray} Exactly four tire-and-wheel assemblies are attached to the structure. & +8 & \checkmark & -- \\
  The wheel assemblies have black tires and yellow wheel centers. & +4 & \checkmark & -- \\
  \rowcolor{rowgray} Snow is present on both the left and right sides of the road. & +7 & \checkmark & -- \\
  Out-of-focus buildings appear on the right side of the background. & +3 & \checkmark & -- \\
  \multicolumn{4}{l}{\textit{Negative criteria (active errors)}} \\
  \rowcolor{rowgray} Impossible/broken geometry in the mailbox, frame, or wheels. & $-6$ & \ding{55}$^\ast$ & \texttt{GEOMETRY\_ARTIFACT} \\
  Extraneous people, vehicles, signs, text, watermarks, or logos. & $-5$ & \ding{55}$^\ast$ & \texttt{HALLUCINATED\_OBSERVATION} \\
  \rowcolor{rowgray} Cloned or unnaturally repeated mailbox/wheel elements. & $-4$ & \checkmark & -- \\
  Garbled, incomplete, or illegible address-sign characters. & $-6$ & \ding{55}$^\ast$ & \texttt{TEXT\_GARBLING} \\
  \rowcolor{rowgray} Inconsistent lighting, shadows, or reflections. & $-3$ & \checkmark & -- \\
  Color bleeding or material confusion between parts. & $-3$ & \checkmark & -- \\
  \bottomrule
\end{tabularx}
\vspace{2pt}
\footnotesize $^\ast$ For negative criteria, \ding{55}\ marks a \emph{triggered} error (the problem
\emph{is} present) -- the opposite polarity from positive criteria, where \ding{55}\ marks something
\emph{missing}. Final score: $\max(0, (44 - 17)/54 \times 100) = 50.0$.
\end{table}

The scoring model's explanations make the failures concrete: the \texttt{MISSING\_STEP} verdict
notes that ``the mailbox does not resemble a car; it is a standard mailbox shape with wheels
underneath,'' the \texttt{GEOMETRY\_ARTIFACT} verdict notes that ``the front green support post
passes straight through the front tire,'' and the \texttt{TEXT\_GARBLING} verdict notes that ``the
small text on the front right edge and the small `RD' are somewhat garbled.''

Table~\ref{tab:example1-compare} then shows how all four systems fared on this same prompt against
their own independently-authored (but semantically equivalent) rubric.

\begin{table}[H]
\centering
\caption{All four systems on the ``Car-shaped Mailbox'' prompt.}
\label{tab:example1-compare}
\small
\begin{tabularx}{\textwidth}{L C P{8.5cm}}
  \toprule
  \textbf{Model} & \textbf{Final score} & \textbf{Notable criteria} \\
  \midrule
  \rowcolor{rowgray} Gemini 3 Pro Image & \textbf{100.0/100} & All positive criteria met; no negative criteria triggered. \\
  Hunyuan 3.0 & \textbf{86.8/100} & All positive criteria met; one \texttt{GEOMETRY\_ARTIFACT} triggered (front wheel clips the support frame). \\
  \rowcolor{rowgray} FLUX.2 & \textbf{50.0/100} & Mailbox does not read as car-shaped (\texttt{MISSING\_STEP}, $+10$ missed); \texttt{GEOMETRY\_ARTIFACT}, \texttt{HALLUCINATED\_OBSERVATION}, and \texttt{TEXT\_GARBLING} all triggered. \\
  Ideogram 3.0 & \textbf{25.5/100} & Four separate positive misses: not car-shaped ($+10$), address sign missing a digit (\texttt{TEXT\_GARBLING}, $+10$), only two tires visible (\texttt{MISCOUNT}, $+8$), no visible frame ($+4$); plus one \texttt{GEOMETRY\_ARTIFACT}. \\
  \bottomrule
\end{tabularx}
\end{table}

Every system correctly rendered the exact address string and the snow/background details -- the
differentiating failure across all four is whether the mailbox actually reads as ``car-shaped''
rather than as a generic mailbox with wheels bolted on, plus how many geometry and hallucination
errors each system introduced on top.

\subsection{Example: ``Swimming Competition''}

\textbf{Prompt:} \textit{``The image is of a swimming competition at an indoor pool, with four young
boy swimmers positioned on starting blocks numbered 5 to 8, preparing to enter the water. Coaches or
officials stand nearby. Spectators sit in the background behind a yellow railing. [...] From an
eye-level angle.''}

This prompt combines exact counting (four swimmers, four blocks) with exact sequential labeling
(blocks numbered 5, 6, 7, 8) -- two of the hardest requirement types in the taxonomy.

\begin{figure}[H]
  \centering
  \IfFileExists{figures/example_swimming.png}{%
    \includegraphics[width=\textwidth]{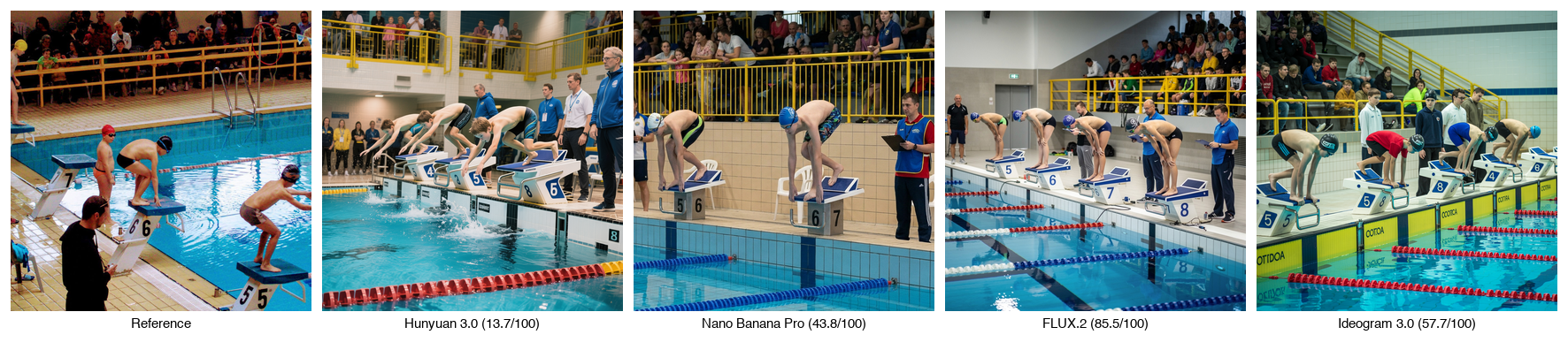}%
  }{%
    \fbox{\parbox[c][3.5cm][c]{0.95\textwidth}{\centering
      Replace this box with \texttt{figures/example\_swimming.png}.}}%
  }
  \caption{Reference photograph and all four generated outputs for this prompt. FLUX.2 is the only
  system that renders a clean 5--6--7--8 block sequence; every other system either miscounts the
  swimmers, garbles the block numbers, or both.}
  \label{fig:example-swimming}
\end{figure}

Hunyuan 3.0 produced the weakest result on this prompt (13.7/100); Table~\ref{tab:example2-rubric}
shows its full rubric and verdicts.

\begin{table}[H]
\centering
\caption{Full rubric and verdicts for Hunyuan 3.0 on the ``Swimming Competition'' prompt (final score 13.7/100).}
\label{tab:example2-rubric}
\small
\begin{tabularx}{\textwidth}{P{8.7cm} C C L}
  \toprule
  \textbf{Criterion} & \textbf{Weight} & \textbf{Verdict} & \textbf{Failure code} \\
  \midrule
  \multicolumn{4}{l}{\textit{Positive criteria (possible weight: 51)}} \\
  \rowcolor{rowgray} Scene is set inside an indoor swimming pool facility. & +8 & \checkmark & -- \\
  Exactly four young boy swimmers are shown. & +10 & \ding{55} & \texttt{MISCOUNT} \\
  \rowcolor{rowgray} Swimmers occupy starting blocks labeled 5, 6, 7, and 8. & +10 & \ding{55} & \texttt{TEXT\_GARBLING} \\
  All four swimmers are shown preparing to dive. & +8 & \ding{55} & \texttt{MISCOUNT} \\
  \rowcolor{rowgray} At least one coach or official stands on the pool deck. & +5 & \checkmark & -- \\
  Spectators are visible behind a yellow railing. & +5 & \checkmark & -- \\
  \rowcolor{rowgray} Camera viewpoint is eye-level, not overhead or underwater. & +3 & \checkmark & -- \\
  White/beige/gray tile surfaces are visible around the pool. & +2 & \checkmark & -- \\
  \multicolumn{4}{l}{\textit{Negative criteria (active errors)}} \\
  \rowcolor{rowgray} Anatomical/geometric artifacts on people (limbs, hands, faces). & $-8$ & \ding{55}$^\ast$ & \texttt{GEOMETRY\_ARTIFACT} \\
  Hallucinated extraneous objects, signs, watermarks, or logos. & $-4$ & \checkmark & -- \\
  \rowcolor{rowgray} Duplicated or cloned swimmers, officials, or blocks. & $-5$ & \checkmark & -- \\
  Illegible or garbled numbers/text on blocks or signage. & $-4$ & \ding{55}$^\ast$ & \texttt{TEXT\_GARBLING} \\
  \rowcolor{rowgray} Implausible lighting, shadows, or splash behavior. & $-4$ & \ding{55}$^\ast$ & \texttt{PHYSICS\_VIOLATION} \\
  Color bleeding or material confusion. & $-4$ & \checkmark & -- \\
  \bottomrule
\end{tabularx}
\vspace{2pt}
\footnotesize $^\ast$ Triggered (present). Final score: $\max(0, (23 - 16)/51 \times 100) = 13.7$.
\end{table}

The scorer's explanations are again concrete: only three swimmers are visible rather than four; the
starting blocks are numbered ``5/4, 2/5, and 8/6'' rather than a clean 5--6--7--8 sequence; and
splashes appear in the water \emph{before} the swimmers reach it, an explicit
\texttt{PHYSICS\_VIOLATION}. Table~\ref{tab:example2-compare} shows the same prompt scored across
all four systems.

\begin{table}[H]
\centering
\caption{All four systems on the ``Swimming Competition'' prompt.}
\label{tab:example2-compare}
\small
\begin{tabularx}{\textwidth}{L C P{8.5cm}}
  \toprule
  \textbf{Model} & \textbf{Final score} & \textbf{Notable criteria} \\
  \midrule
  \rowcolor{rowgray} FLUX.2 & \textbf{85.5/100} & All positive criteria met, including the exact 5--6--7--8 block sequence; only one \texttt{GEOMETRY\_ARTIFACT} triggered (officials' hands). \\
  Ideogram 3.0 & \textbf{57.7/100} & Block numbering garbled (\texttt{TEXT\_GARBLING}, $+9$ missed); \texttt{GEOMETRY\_ARTIFACT} and a second \texttt{TEXT\_GARBLING} (poolside signage) triggered. \\
  \rowcolor{rowgray} Gemini 3 Pro Image & \textbf{43.8/100} & Correct count of four swimmers, but block numbering garbled ($+8$ missed); extra hallucinated swimmers on deck, plus \texttt{GEOMETRY\_ARTIFACT} and two \texttt{TEXT\_GARBLING} triggers. \\
  Hunyuan 3.0 & \textbf{13.7/100} & Only three of four swimmers rendered (\texttt{MISCOUNT} on two separate criteria, $+10$ and $+8$); block numbering garbled ($+10$); \texttt{GEOMETRY\_ARTIFACT}, \texttt{TEXT\_GARBLING}, and \texttt{PHYSICS\_VIOLATION} all triggered. \\
  \bottomrule
\end{tabularx}
\end{table}

This prompt illustrates a pattern visible across the full 48-prompt set: exact sequential text
labeling (``5, 6, 7, 8'') is the single hardest requirement type for every system -- even FLUX.2,
the only model to satisfy it here, does so alongside a geometry artifact elsewhere in the frame --
and undercounting people is the failure mode most strongly correlated with a model's weakest
scores.

\section{Results}
\label{sec:results}

Table~\ref{tab:main-results} reports the matched-N ranking across all four systems on the
identical set of 48 prompts.

\begin{table}[H]
\centering
\caption{Matched-N ranking: identical 48 prompts, all four models. Net score is the mean unnormalized weighted balance (earned $-$ incurred); unmet criteria and active errors are mean counts per prompt.}
\label{tab:main-results}
\small
\begin{tabularx}{\textwidth}{L C C C C}
  \toprule
  \textbf{Model} & \textbf{Final score} & \textbf{Net score} & \textbf{Avg.\ unmet criteria} & \textbf{Avg.\ active errors} \\
  \midrule
  \rowcolor{rowgray} Gemini 3 Pro Image (Nano Banana Pro) & \textbf{84.8/100} & +44.3 & 0.5 & 0.8 \\
  Black Forest Labs FLUX.2 & \textbf{82.3/100} & +42.6 & 0.5 & 1.0 \\
  \rowcolor{rowgray} Ideogram 3.0 & \textbf{65.7/100} & +34.1 & 1.4 & 1.5 \\
  Hunyuan 3.0 & \textbf{63.3/100} & +33.3 & 1.4 & 1.7 \\
  \bottomrule
\end{tabularx}
\end{table}

Gemini 3 Pro Image and FLUX.2 form a clear leading tier, separated from each other by only 2.5
points, while Ideogram 3.0 and Hunyuan 3.0 form a trailing tier roughly 17--21 points lower. The
gap between tiers is far larger than the gap within either tier, indicating a real capability
separation on this hard-prompt subset rather than noise in the grading process.

\begin{figure}[H]
  \centering
  \IfFileExists{figures/qualitative_comparison.png}{%
    \includegraphics[width=\textwidth]{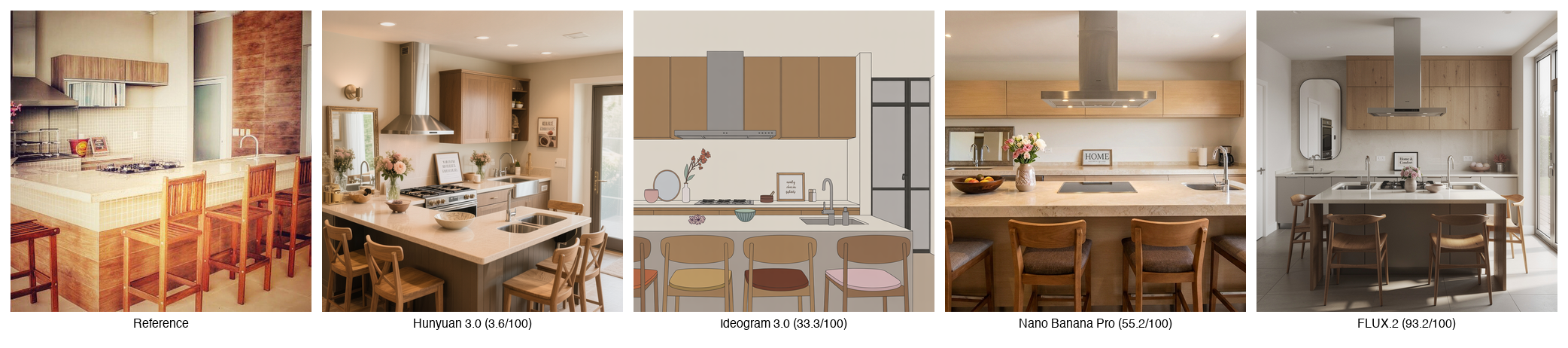}%
  }{%
    \fbox{\parbox[c][4cm][c]{0.95\textwidth}{\centering
      Replace this box with \texttt{figures/qualitative\_comparison.png}.}}%
  }
  \caption{Reference photograph and all four generated outputs for the prompt ``Modern Kitchen'',
  which produced the widest score spread of any prompt in the evaluation set
  (89.6-point range between the best- and worst-scoring model). The prompt requires exactly four
  chairs, two stools, and a specific countertop material -- FLUX.2 (93.2/100) satisfies nearly all
  positive criteria, while Hunyuan 3.0 (3.6/100) miscounts most requested objects and introduces
  several geometric artifacts.}
  \label{fig:main-result}
\end{figure}

\subsection{Failure-Code Analysis}

Table~\ref{tab:failure-codes} breaks each model's errors down by failure code, separating
\emph{positive-axis} failures (requested elements the model failed to deliver) from
\emph{negative-axis} failures (errors the model introduced on its own initiative).

\begin{table}[H]
\centering
\caption{Top failure codes per model on the matched-48 set (occurrence counts across all 48 prompts). Codes with fewer than 2 combined occurrences across all models are omitted for brevity.}
\label{tab:failure-codes}
\small
\begin{tabularx}{\textwidth}{L L L}
  \toprule
  \textbf{Model} & \textbf{Positive-axis (missed requests)} & \textbf{Negative-axis (active errors)} \\
  \midrule
  \rowcolor{rowgray} Gemini 3 Pro Image & \texttt{MISCOUNT} (9), \texttt{SPATIAL\_RELATION\_ERROR} (7) & \texttt{TEXT\_GARBLING} (16), \texttt{GEOMETRY\_ARTIFACT} (12), \texttt{HALLUCINATED\_OBSERVATION} (9) \\
  FLUX.2 & \texttt{MISCOUNT} (8), \texttt{MISSING\_STEP} (4) & \texttt{TEXT\_GARBLING} (19), \texttt{GEOMETRY\_ARTIFACT} (16), \texttt{PHYSICS\_VIOLATION} (7) \\
  \rowcolor{rowgray} Ideogram 3.0 & \texttt{MISSING\_STEP} (24), \texttt{TEXT\_GARBLING} (12) & \texttt{TEXT\_GARBLING} (27), \texttt{GEOMETRY\_ARTIFACT} (24), \texttt{HALLUCINATED\_OBSERVATION} (12) \\
  Hunyuan 3.0 & \texttt{TEXT\_GARBLING} (20), \texttt{MISCOUNT} (12) & \texttt{TEXT\_GARBLING} (29), \texttt{GEOMETRY\_ARTIFACT} (27), \texttt{HALLUCINATED\_OBSERVATION} (15) \\
  \bottomrule
\end{tabularx}
\end{table}

Two qualitatively distinct failure patterns emerge. The leading tier (Gemini 3 Pro Image, FLUX.2)
loses points mainly to \texttt{MISCOUNT} -- generating the right kind of object but the wrong
number of instances -- alongside a moderate rate of geometric artifacts. The trailing tier
(Ideogram 3.0, Hunyuan 3.0) shows a materially different profile:
\texttt{TEXT\_GARBLING} dominate the positive axis for both, and Ideogram 3.0 adds a high \texttt{MISSING\_STEP} rate, meaning these models more often fail to render legible text and, for Ideogram 3.0, fail to attempt a requested element at all, rather than attempting it and getting a detail wrong. \texttt{TEXT\_GARBLING} and \texttt{GEOMETRY\_ARTIFACT} are the two most common
negative-axis errors for every model in the study, indicating that legible-text rendering and
clean geometry remain unsolved problems even for the strongest systems evaluated here.

\section{Discussion}
\label{sec:discussion}

The 21.5-point gap between the strongest model (Gemini 3 Pro Image, 84.8/100) and the weakest
(Hunyuan 3.0, 63.3/100) on identical, maximally hard prompts is substantially larger than gaps
typically reported on average-case T2I benchmarks, consistent with the motivation in
Section~\ref{sec:introduction}: difficulty-targeted prompt selection surfaces capability
differences that average-case evaluation compresses. Because the model that authors each rubric is
never the model that renders its verdicts (Section~4.3), the ranking is unlikely to simply reflect
one judge's stylistic preferences masquerading as genuine output-quality differences.

The failure-code breakdown is arguably more actionable for model selection than the aggregate score
alone: a team whose product requires legible in-image text should weight \texttt{TEXT\_GARBLING}
rates heavily regardless of aggregate rank, while a team generating scenes with many discrete
objects should weight \texttt{MISCOUNT} and \texttt{MISSING\_STEP} rates. On this dataset, the two
leading models are broadly interchangeable in aggregate score but diverge somewhat in their error
profile (FLUX.2 shows a higher \texttt{PHYSICS\_VIOLATION} rate; Gemini 3 Pro Image shows a higher
\texttt{HALLUCINATED\_OBSERVATION} rate at a similar overall active-error count), which a single
top-line number would not reveal.

\subsection{Limitations}

This study evaluates four candidate systems from a broader field of five originally considered; one
additional candidate was excluded from the final comparison and is not reported here. Complexity
scoring is performed by a single LLM classifier rather than by human annotators, so the notion of
``hardest'' reflects that classifier's judgment of difficulty rather than a ground-truth human
difficulty ranking, though the qualitative hardness factors it surfaces (legible text, multi-object
counting, spatial constraints) align with well-established compositional failure modes in the T2I
literature \cite{huang2023t2icompbench, huang2025t2icompbenchpp}. The rubric-scoring model (Gemini
3.1 Pro Preview) is from the same model family as one of the four candidate generation systems
(Gemini 3 Pro Image); while rubric authorship is handled by a separate, independent model
(GPT-5.4-Pro) specifically to mitigate self-grading bias, a same-family scorer is a residual risk
worth flagging.

A second, related risk comes from the rubric writer seeing each system's actual generated image
while authoring criteria (Section~4.3.1): although the writer never renders a met/triggered verdict
itself, it does decide which criteria to include and how to phrase them with that specific image in
view, which is a weaker independence guarantee than a rubric written purely from the prompt before
any image exists. We mitigate this by having a separate model apply the final verdict, but cannot
fully rule out the writer subtly calibrating criteria difficulty to what it observes. Relatedly,
every system was queried with its provider's default generation settings (Appendix~\ref{app:gen-params})
and no fixed random seed, so results reflect default, single-sample, out-of-the-box quality rather
than each system's best achievable output under prompt engineering or parameter tuning -- a
different comparison protocol could shift absolute (though probably not relative) rankings. A
human-SME calibration pass against a stratified subset of scores is supported by the evaluation
pipeline but was not run for this study.

\subsection{Future Work}

Natural extensions include: (1) completing a human-SME calibration pass to quantify agreement
between the automated judge and human raters; (2) extending the matched-N comparison to the fifth
candidate model excluded here; (3) repeating the complexity-scoring pass with an ensemble of
classifiers to test the robustness of the ``hardest 48'' selection; and (4) a targeted follow-up
study isolating text-rendering and object-counting prompts specifically, given how much of the
score gap in this study is attributable to those two failure codes.

\section{Conclusion}
\label{sec:conclusion}

On the 48 hardest prompts in the DataSeeds.AI DSD dataset, graded by an independent-judge rubric
methodology that separates rubric authorship from rubric scoring, Gemini 3 Pro Image and FLUX.2
form a clear leading tier (84.8 and 82.3 out of 100, respectively), well ahead of Ideogram 3.0 and
Hunyuan 3.0 (65.7 and 63.3). The leading and trailing tiers fail in qualitatively different ways --
miscounted objects and geometric artifacts for the leaders, versus omitted requested elements and
garbled text for the trailing models -- information that a single aggregate score would not
surface, and that should directly inform model selection for compositionally demanding
image-generation use cases.

\section*{Data and Code Availability}

The 48 selected prompts, reference images, per-model generated images, full independent-judge
rubrics, per-criterion verdicts, and the evaluation pipeline code are retained internally at
\organizationname\ and available on request from the corresponding author.

\section*{Ethics Statement}

All images were generated from publicly available dataset prompts (DataSeeds.AI DSD) using paid,
provider-hosted generation APIs, with no personal or sensitive data involved. Some generation
requests were declined by a provider's own content-moderation system prior to reaching our
evaluation pipeline; those prompts were excluded from the matched comparison rather than retried
or circumvented.

\bibliographystyle{plain}
\bibliography{template_references}

\appendix

\section{Supplementary Material}
\label{app:supplementary}

\subsection{Additional Qualitative Examples}
\label{app:gallery}

The two worked examples in Section~\ref{sec:worked-examples} were chosen to illustrate the rubric
mechanism in detail; this appendix adds four further prompts spanning different hardness factors
from Table~\ref{tab:complexity-rubric} (legible text on a monument, multi-object spatial
arrangement, physics/reflection constraints, and nested color-attribute binding) to give a broader
sense of how failure patterns vary by prompt type across the 48-prompt evaluation set.

\begin{figure}[H]
  \centering
  \IfFileExists{figures/appendix_monument.png}{%
    \includegraphics[width=\textwidth]{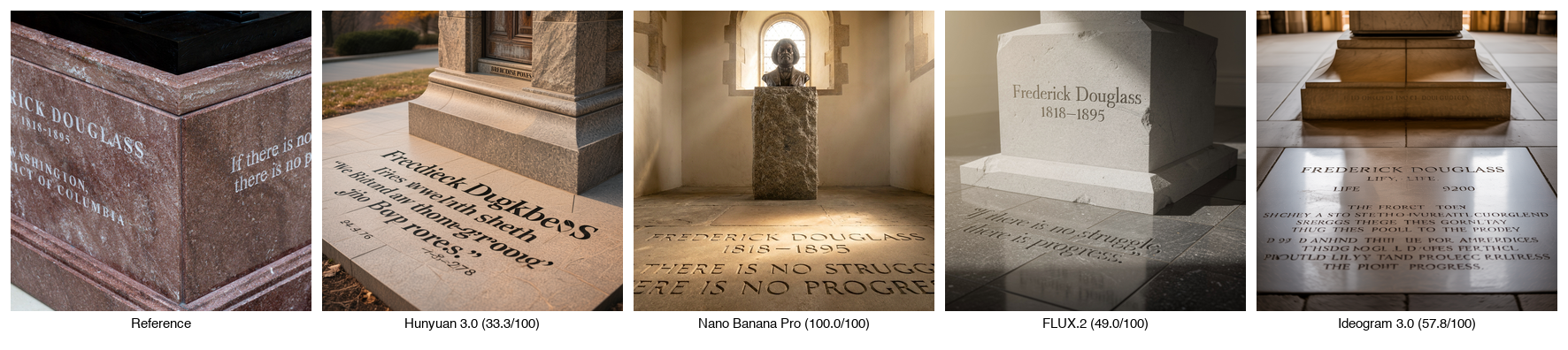}%
  }{%
    \fbox{\parbox[c][3.5cm][c]{0.95\textwidth}{\centering
      Replace this box with \texttt{figures/appendix\_monument.png}.}}%
  }
  \caption{``Legacy of Timeless Words'' -- legible-text and semantic-binding hardness
  factors (an engraved stone monument with a specific name, date, and quotation). Gemini 3 Pro Image
  is the only system with a perfect score here; the other three each fail differently -- Hunyuan 3.0
  renders all engraved text as unreadable glyphs, FLUX.2 engraves the correct name and dates on the
  pedestal rather than the floor as specified, and Ideogram 3.0 substitutes the requested life dates
  with an unrelated number.}
  \label{fig:appendix-monument}
\end{figure}

\begin{figure}[H]
  \centering
  \IfFileExists{figures/appendix_houses.png}{%
    \includegraphics[width=\textwidth]{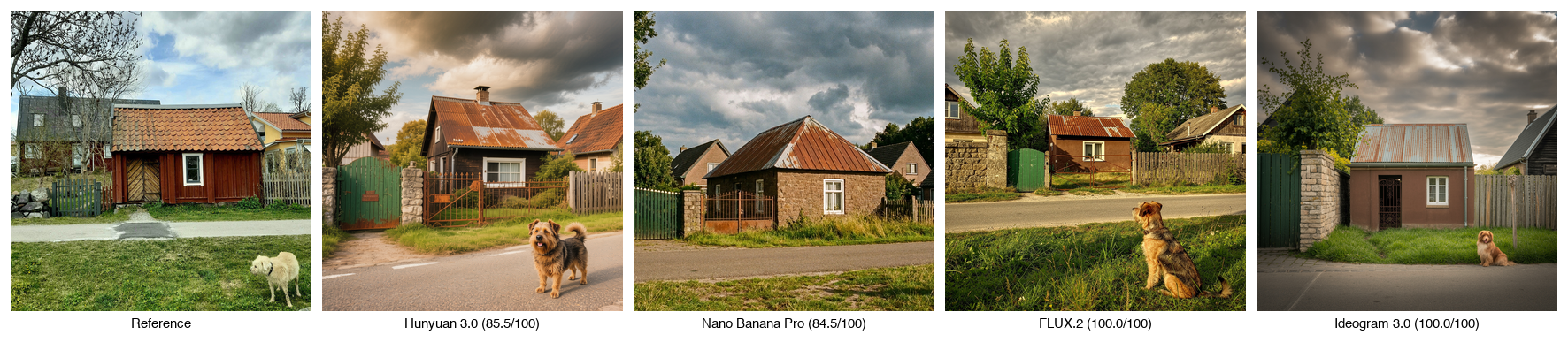}%
  }{%
    \fbox{\parbox[c][3.5cm][c]{0.95\textwidth}{\centering
      Replace this box with \texttt{figures/appendix\_houses.png}.}}%
  }
  \caption{``3 houses under the clouds'' -- precise spatial arrangement and multiple
  distinct object classes (three named buildings, a specific dog breed, specific gate and roof
  materials). This is one of the few prompts in the set where FLUX.2 and Ideogram 3.0 both reach a
  perfect score while Gemini 3 Pro Image and Hunyuan 3.0 lose a small amount of credit, illustrating
  that the leading tier's advantage is not uniform across every prompt.}
  \label{fig:appendix-houses}
\end{figure}

\begin{figure}[H]
  \centering
  \IfFileExists{figures/appendix_snails.png}{%
    \includegraphics[width=\textwidth]{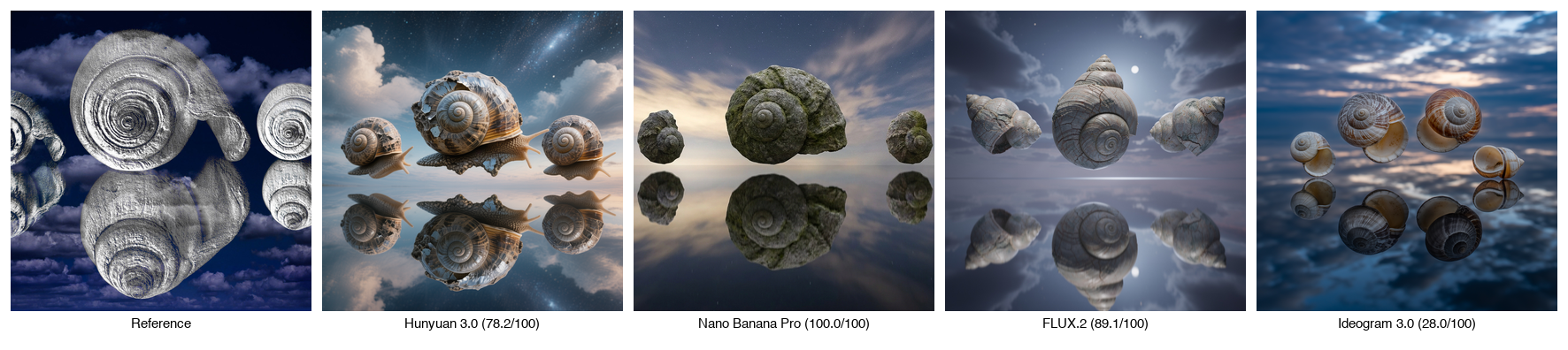}%
  }{%
    \fbox{\parbox[c][3.5cm][c]{0.95\textwidth}{\centering
      Replace this box with \texttt{figures/appendix\_snails.png}.}}%
  }
  \caption{``Surreal Spiral Dreams'' -- a perfect-reflection physics constraint (three
  snail shells that must appear as six via mirror reflection) combined with precise counting.
  Ideogram 3.0 scores far below the other three systems here (28.0/100), driven by
  \texttt{PHYSICS\_VIOLATION} and \texttt{MISCOUNT} verdicts on the reflection requirement
  specifically.}
  \label{fig:appendix-snails}
\end{figure}

\begin{figure}[H]
  \centering
  \IfFileExists{figures/appendix_balcony.png}{%
    \includegraphics[width=\textwidth]{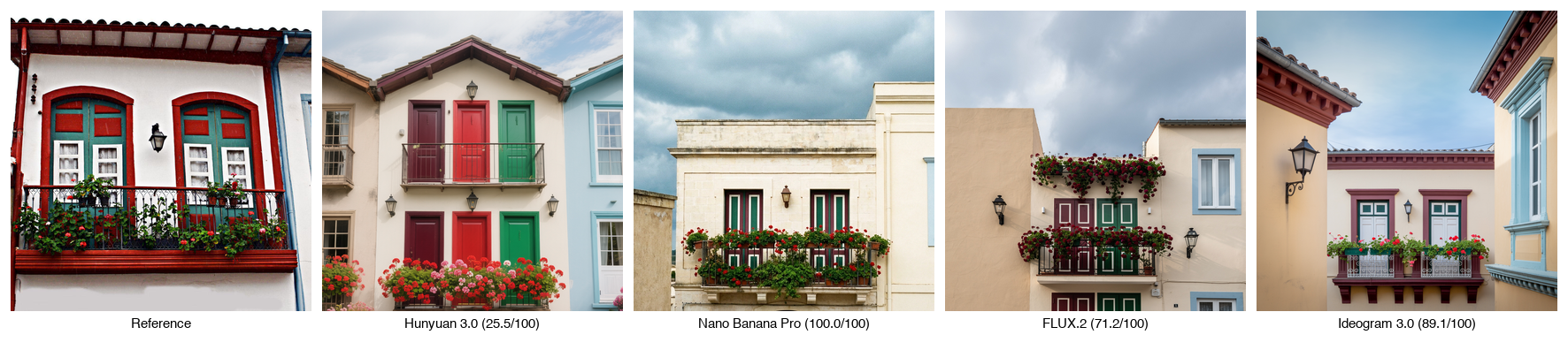}%
  }{%
    \fbox{\parbox[c][3.5cm][c]{0.95\textwidth}{\centering
      Replace this box with \texttt{figures/appendix\_balcony.png}.}}%
  }
  \caption{``Colourful Balcony with geranium flowers'' -- nested attribute binding across
  three buildings, each with its own wall color, lantern, and balcony detail. Hunyuan 3.0 scores
  lowest (25.5/100) here, misattributing the right building's wall color and window-frame color and
  miscounting the middle balcony's doors, on top of a duplicated ground-floor geometry artifact.}
  \label{fig:appendix-balcony}
\end{figure}

\subsection{Generation Parameters}
\label{app:gen-params}

Table~\ref{tab:gen-params} lists every parameter explicitly set for each candidate system's image
generation API call. No system was given a fixed random seed, so each generation is a single,
non-reproducible sample rather than a best-of-$k$ selection; no manual prompt engineering, negative
prompting, or per-system parameter tuning was performed beyond what is listed. Every parameter not
listed (sampling steps, guidance/CFG scale, style presets, safety-filter strictness, and so on) was
left at that provider's own default for the endpoint used. This is a deliberate simplification: the
comparison reflects each system's out-of-the-box default behavior on identical prompts, not each
system's best achievable quality under bespoke tuning (see Limitations).

\begin{table}[H]
\centering
\caption{Explicit generation parameters passed to each system's API. All other parameters (sampling steps, guidance scale, style, safety filtering, etc.) were left at the provider's default.}
\label{tab:gen-params}
\small
\begin{tabularx}{\textwidth}{L L L}
  \toprule
  \textbf{System} & \textbf{Endpoint} & \textbf{Explicit parameters} \\
  \midrule
  \rowcolor{rowgray} Hunyuan 3.0 & Pixazo \modelname{hunyuan-image} generate & \texttt{image\_size=square\_hd}, \texttt{num\_images=1} \\
  Gemini 3 Pro Image & Google GenAI \modelname{generate\_content} (\modelname{gemini-3-pro-image}) & \texttt{response\_modalities=[IMAGE]} \\
  \rowcolor{rowgray} FLUX.2 & Black Forest Labs \modelname{flux-2-pro-preview} & \texttt{width=1024}, \texttt{height=1024} \\
  Ideogram 3.0 & Ideogram \modelname{ideogram-v3/generate} & \texttt{aspect\_ratio=1x1} \\
  \bottomrule
\end{tabularx}
\end{table}

\subsection{Rubric Authoring Prompt (Excerpt)}

The writer model is instructed to produce criteria that are \emph{atomic} (one evaluable claim per
criterion), \emph{specific} (binary, objective, $\geq$75\% expected inter-rater agreement),
\emph{self-contained} (expected values stated inline, e.g. ``shows exactly 3 balloons'' rather than
``shows the correct number of balloons''), \emph{MECE and non-redundant} (no criterion that merely
summarizes others), and \emph{weighted by Core vs.\ Additional} importance (essential requirements
weighted higher than nice-to-have enhancements).

\end{document}